\documentclass{article}
\usepackage{wrapfig}
\usepackage{float} 
\usepackage{tensor}
\usepackage{tabu}
\usepackage{algorithmicx}
\usepackage{algorithm}
\usepackage{mathtools}
\usepackage{booktabs}
\usepackage[colorlinks,citecolor=red,urlcolor=blue,bookmarks=false,hypertexnames=true]{hyperref}
\hypersetup{
    colorlinks,
    citecolor=black,
    filecolor=black,
    linkcolor=black,
    urlcolor=black
}
\usepackage{amssymb}
\usepackage{bbm}
\usepackage{graphicx}
\usepackage{array,etoolbox}
\usepackage{dsfont}
\usepackage{enumitem,kantlipsum}

\preto\tabular{\setcounter{magicrownumbers}{0}}
\newcounter{magicrownumbers}

\DeclarePairedDelimiterX{\infdivx}[2]{\big(}{\big)}{%
  #1\;\delimsize\|\;#2%
}
\usepackage{subfig}
\usepackage[final]{corl_2020}
\usepackage{multicol}
\usepackage{multirow}
\usepackage{amsmath}
\makeatletter

\usepackage{titlesec}
\titlespacing\section{0pt}{0pt plus 2pt minus 2pt}{0pt plus 2pt minus 2pt}
\titlespacing\subsection{0pt}{3pt plus 4pt minus 2pt}{0pt plus 2pt minus 2pt}
\titlespacing\subsubsection{0pt}{3pt plus 4pt minus 2pt}{0pt plus 2pt minus 2pt}

\DeclareMathOperator*{\argmax}{arg\,max}

\newcommand*{\inlineequation}[2][]{%
  \begingroup
    \refstepcounter{equation}%
    \ifx\\#1\\%
    \else
      \label{#1}%
    \fi
    \relpenalty=10000 %
    \binoppenalty=10000 %
    \ensuremath{%
      #2%
    }%
    ~\@eqnnum
  \endgroup
}
\let\Right\right
\let\Left\left
\def\right#1{\Right#1\@ifnextchar){\!\right}{}}
\def\left#1{\Left#1\@ifnextchar({\!\left}{}}
\makeatother
\title{Example-Driven Model-Based Reinforcement Learning for Solving Long-Horizon Visuomotor Tasks}
\newcommand{\acronym}{\operatorname{\text{EMBER}}}
\newcommand{\enc}{f_\text{enc}}
\newcommand{\dec}{f_\text{dec}}
\newcommand{\vae}{f_\text{vae}}
\newcommand{\rew}{f^k_{\mathcal{R}}}
\newcommand{\q}{f^k_{Q}}
\newcommand{\dyn}{f_{\mathcal{T}}}
\usepackage[noend]{algpseudocode}
\author{
  Bohan Wu, Suraj Nair, Li Fei-Fei\footnotemark[2]\hspace{1.5mm}, Chelsea Finn\footnotemark[2]\hspace{1.5mm}\\
  Stanford University, Stanford, CA\\
  \tt\small \texttt{\{bohanwu, surajn, feifeili, cbfinn\}@cs.stanford.edu}
}
\usepackage[toc,page,header]{appendix}
\usepackage{minitoc}

\begin{document}
\doparttoc
\faketableofcontents
\parttoc
\maketitle

\definecolor{rc}{RGB}{0,0,0}

\begin{abstract}
\textcolor{rc}{In this paper, we study the problem of learning a repertoire of low-level skills from raw images that can be sequenced to complete long-horizon visuomotor tasks.} Reinforcement learning (RL) is a promising approach for acquiring short-horizon skills autonomously. However, the focus of RL algorithms has largely been on the success of those individual skills, \textcolor{rc}{more so than learning and grounding a large repertoire of skills that can be sequenced to complete extended multi-stage tasks}. The latter demands robustness and persistence, as errors in skills can compound over time, and may require the robot to have a number of primitive skills in its repertoire, rather than just one. To this end, we introduce $\acronym$, a model-based RL method for \textcolor{rc}{learning primitive skills that are suitable for completing long-horizon visuomotor tasks}. $\acronym$ learns and plans using a learned model, critic, and success classifier, where the success classifier serves both as a reward function for RL and as a grounding mechanism to continuously detect if the robot should retry a skill when unsuccessful or under perturbations. Further, the learned model is task-agnostic and trained using data from all skills, enabling the robot to efficiently learn a number of distinct primitives. \textcolor{rc}{These visuomotor primitive skills and their associated pre- and post-conditions can then be directly combined with off-the-shelf symbolic planners to complete long-horizon tasks}. On a Franka Emika robot arm, we find that $\acronym$ enables the robot to complete three long-horizon visuomotor tasks at 85\% success rate, such as organizing a desk, a cabinet, and drawers, which require sequencing up to 12 skills, involve 14 unique learned primitives, and demand generalization to novel objects. 
\end{abstract}


\section{Introduction}
We want robots to robustly complete a variety of long-horizon visuomotor manipulation tasks. Such tasks can be completed by composing a sequence of low-level primitive skills, otherwise known as ``manipulation primitives''~\cite{kroger2010manipulation} or ``options''~\cite{bacon2017option}. 
For example, the long-horizon task of ``organizing an office desk'' often means performing a long sequence of shorter tasks, such as picking objects up, opening or closing drawers, placing objects into drawers, and putting pens into pen-holders.
However, completing such a long-horizon task may require the robot to learn many low-level visuomotor skills, compose them sequentially, and generalize across novel objects. 
During task execution, the robot also needs to detect its own failures (e.g. accidentally dropping an object) and correct them \begin{figure}[h]
\vspace{-5px}
    \centering
    \includegraphics[width=\textwidth]{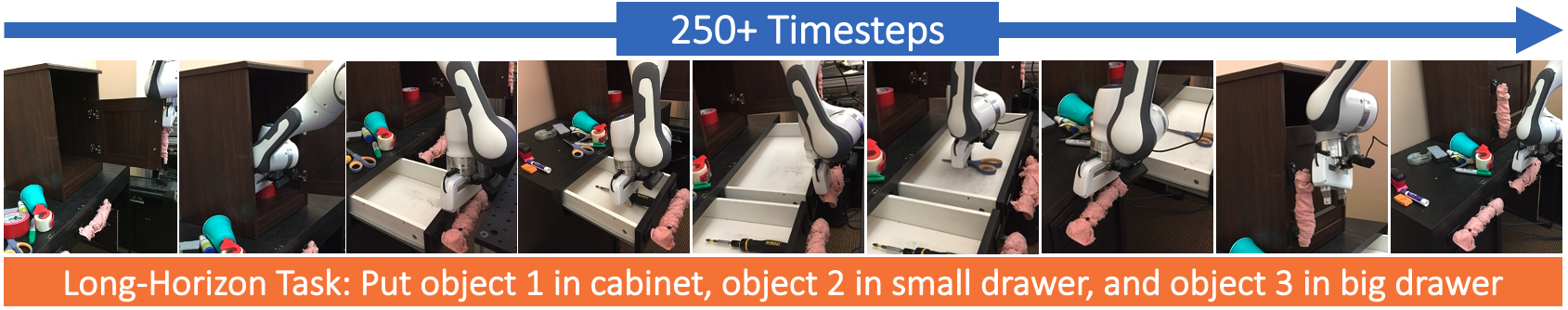}
    \caption{\small \textbf{Example-Driven Model-Based RL ($\acronym$).} $\acronym$ enables a Franka Emika robot to complete long-horizon manipulation tasks by sequencing up to 12 unique skills at 85\% success with novel objects from raw image observations. Here, the robot organizes a desk, which involves a sequence of 10 unique skills.}
    \label{fig:pull}
    \vspace{-10px}
\end{figure}(e.g. picking the object up again) before continuing to the remaining parts of the task.
We hope to endow robots with such generalization and robustness. 

Approaches such as symbolic planning~\cite{smolensky1987connectionist}, hierarchical RL~\cite{Sutton99betweenmdps, Barto:2003:RAH:608557.608576, DBLP:journals/corr/BaconHP16, DBLP:journals/corr/abs-1805-08296nachum, levy2018hierarchical}, and hierarchical planning~\cite{Konidaris2018FromST} can accomplish compositional task generalization by reasoning in an abstract or symbolic space. While these approaches have shown promising results on tasks such as simulated tool use~\cite{Toussaint2018DifferentiablePA} and non-vision-based manipulation~\cite{Konidaris2018FromST},
they rely on the ability to acquire a repertoire of robust, low-level motor skills. These low-level skills can in principle be acquired through reinforcement learning, but a number of challenges remain. First, the acquisition of robust and persistent low-level visuomotor manipulation skills is still challenging, and the errors in completing any sub-sequence of a long-horizon task can compound over time and lead to overall task failure. Second, acquiring the requisite low-level visuomotor skills requires data-efficient learning and reward or task specification across a broad set of skills. \textcolor{rc}{Finally, sequencing the learned primitive skills to perform a long-horizon task requires persistent grounding of the robot's visual observation within the broader abstract plan}.

\textcolor{rc}{To address these challenges, we introduce example-driven model-based RL ($\acronym$), an algorithm that learns a dynamics model and a set of Q-functions and success classifiers to acquire a repertoire of visuomotor skills that are both robust and can be \textcolor{rc}{directly} grounded into symbols for symbolic planning of a long-horizon visuomotor task}. More specifically, the success classifiers in $\acronym$ enable both robustness and skill grounding, as they allow the robot to determine the pre- and post-conditions of a skill. For example, a reward function and post-condition for the skill of opening a drawer can be obtained from learning a binary image classifier of ``whether the drawer is open'' given the robot's pixel observation. These success classifiers require a modest number of example images, making it easy to specify a variety of skills. Finally, by learning a task-agnostic dynamics model, $\acronym$ allows data to be shared across all tasks, leading to greater data-efficiency.

\textcolor{rc}{The main contribution of this work is a framework for learning grounded visuomotor skills that can be directly sequenced by off-the-shelf symbolic planners to complete long-horizon tasks}. Experiments on a Franka Emika Panda robot arm find that $\acronym$ enables the robot to complete three long-horizon tasks from raw pixels with $85\%$ success, such as organizing a desk, a cabinet, and two drawers, which require up to 14 learned unique primitives, sequencing up to 12 skills, and generalization to novel objects. 
Moreover, our ablation studies confirm that all components of our method including the dynamics model, success classifiers, Q-functions, and stages are essential to performance for tackling long-horizon tasks.

\section{Related Work}
Reinforcement learning (RL) has a rich history of being used for robotic control problems on tasks from locomotion \cite{BENBRAHIM1997283, Kohl2004PolicyGR, Tedrake2004StochasticPG} to object manipulation \cite{GULLAPALLI1995237, PETERS2008682, Deisenroth-RSS-11}. In this work, we focus on learning \emph{vision-based} robotic manipulation skills, a problem studied by many prior works \cite{riedmiller2012, endtoendvisuomotor, Ghadirzadeh2017DeepPP, kalashnikov2018qt}. Model-free deep RL has been effective on short-horizon skills like object grasping \cite{Pinto2016SupersizingSL, kalashnikov2018qt, zeng2018learning}, pushing and throwing
\cite{Ghadirzadeh2017DeepPP, nair2018visual, singh2019end}, and multi-task learning \cite{nair2018visual, Kalashnikov2021MTOptCM} from images. Alternatively, model-based approaches~\cite{pmlr-v97-hafner19a, hafner2019dream}, which explicitly learn the environment forward dynamics from images have also been employed for multi-task vision-based robotic manipulation tasks, either with planning algorithms~\cite{finn2016unsupervised, finn2017deep, ebert2018visual, ebert2018robustness, Lin2019ExperienceEmbeddedVF, Suh2020TheSE, tian2020model,bhardwaj2020information} or for optimizing a parametric policy \cite{rafailov2020offline}. 
Unlike these prior works, our algorithm uses a model in conjunction with task-specific success classifiers and Q-functions
to learn a wide range of skills suitable for sequencing into long-horizon tasks. In Section~\ref{exps}, we find that both components of our approach are essential for robust performance.

Even with powerful reinforcement learning methods, task specification on real robots remains challenging, as engineering rewards on physical systems can be costly and time-consuming \cite{Zhu2020TheIO}. 
Motivated by this, many works have studied the reward specification, including inverse reinforcement learning \cite{apprenticeship_abbeel} with robot demonstrations \cite{ratcliff_maxmargin, ziebart2008maximum, wulfmeier2016maximum, finn2016guided, fu2018learning}, learning rewards from user preferences \cite{sadigh2017active, brown2019deep, palan2019learning, brown2020safe}, and learning rewards from videos of humans \cite{shao2020concept, chen2021learning}. One common approach in visual RL is to learn to reach a goal image using a coarse measure of reward like negative $\ell_2$ pixel distance \cite{ebert2018visual, nair2019hierarchical, wu2021greedy} or temporal distance \cite{savinov2018semiparametric, eysenbach2019search, tian2020model}.
While these techniques have had some success on real robots, such rewards often provide a sparse and difficult to optimize reward signal. Alternatively, a number of recent works learn a classifier from a modest number of examples goal states \cite{Xie2018FewShotGI, fu2018variational, singh2019end, chen2020batch, eysenbach2021replacing, Kalashnikov2021MTOptCM}
and tries to learn agents which maximize the classifier score.
\textcolor{rc}{In this work, we extend this approach to multiple stages of human-provided goals, which we find enables more effective learning of hard exploration skills, and find that such reward specification can be used to learn 14 unique visuomotor skills on a real robot without any demonstrations.}

Like many prior works, the goal of this work is to complete challenging long-horizon visuomotor tasks.
One class of prior work is hierarchical RL, which aims to learn temporally-extended primitive skills and a high-level policy over them \cite{Sutton99betweenmdps, Barto:2003:RAH:608557.608576, DBLP:journals/corr/BaconHP16, DBLP:journals/corr/abs-1805-08296nachum, levy2018hierarchical, Jiang2019LanguageAA}. Such approaches include jointly learning the primitive skills with the high-level policy through goal generation or an ``options'' framework \cite{DBLP:journals/corr/BaconHP16, levy2018hierarchical,bagaria2019option},
learning primitives through intrinsic motivation or other auxiliary objectives \cite{diayn, sharma2020dynamicsaware, Biza2021ActionPF}, from demonstration behavior \cite{ddco, fox2018parametrized,Yu2018OneShotHI, Gupta2019RelayPL, Mandlekar2020LearningTG}, or from low-level, potentially goal conditioned, reward functions \cite{DBLP:journals/corr/abs-1805-08296nachum, Jiang2019LanguageAA, chen2020skewexplore, Hafner2020TowardsGA}. \textcolor{rc}{$\acronym$ is not a hierarchical RL algorithm, as it only learns low-level skills provided to the high-level planner and does not learn a model over options.} Nevertheless, our work is similar to the last group, except that we learn visuomotor primitives on a real robot from rewards derived from human examples. Other works have also explored learning hierarchical policies on top of hand-designed skills \cite{xu2018neural, huang2019neural, Janner2019ReasoningAP}.
These methods are complementary to $\acronym$, as skills learned via $\acronym$ can be incorporated into most hierarchical policy learning algorithm. 

Alternatively, a number of recent works have studied long horizon visual planning~\cite{savinov2018semiparametric, eysenbach2019search, causalinfogan, wangvisualplan, jayaraman2018timeagnostic, nair2019hierarchical, liu2020hallucinative, ichter2020broadlyexploring, pertsch2020longhorizon}, often by combining a structured search approach with learned components like generative models~\cite{causalinfogan, nair2019hierarchical, liu2020hallucinative} or distance functions~\cite{eysenbach2019search, tian2020model}. While these approaches have demonstrated impressive results on problems such as visual navigation~\cite{savinov2018semiparametric, eysenbach2019search}, unlike prior work we learn manipulation tasks on a \emph{real robots} that consists of up to 12 unique skills and 250+ timesteps.

Lastly, the task and motion planning (TAMP) literature \cite{tampinthenow, srivastava_tamp} has extensively studied long-horizon robotic tasks. Unlike these works, we do not assume pre-defined state representations or accurate state estimators, making it possible to handle novel objects in cluttered scenes. Like our work, a number of works have also explored combining learned models and skills with hand-designed or symbolic planning \cite{Toussaint2018DifferentiablePA, Konidaris2018FromST, OpenAI2019SolvingRC} and learning to ground skills for planning \cite{Kaelbling2017LearningCM}.
Unlike these techniques, a key insight in this work is that we can jointly learn visuomotor skills and the associated grounding of their pre- and post-conditions, allowing us to leverage off-the-shelf symbolic planners to execute long horizon tasks in a closed-loop fashion directly from pixels on a real robot. 

\section{Preliminaries}
\noindent \textbf{Tasks and skills}. 
\textcolor{rc}{We consider the problem of learning a repertoire of skills that can be sequenced to complete a long-horizon task $\mathcal{M}$.} We model the robot's environment as a controlled Markov process $\mathcal{E} = \langle \mathcal{S}, \rho_0, \mathcal{A}, \mathcal{T}, \gamma, H \rangle$, with an image observation space $s \in \mathcal{S}$, an initial state distribution $\rho_0$, an action space $a \in \mathcal{A}$, a dynamics model $\mathcal{T}: \mathcal{S} \times \mathcal{A} \times \mathcal{S} \to \mathbb{R}$, a discount factor $\gamma \in [0, 1)$, and a finite horizon $H$. We notice that many long-horizon tasks are composed of a sequence of lower-level skills. For example, to organize an office desk, the robot needs to acquire a number of skills such as grasping an object, opening a drawer, and placing an object in a drawer. Let $K$ denote the total number of unique skills the robot has acquired, and $k \in [1, K]$ denote the $k^{th}$ skill in the robot's skill repertoire. Depending on the environment state, the order in which the skills should be executed to successfully complete a long-horizon task varies. For example, to organize a desk, the robot may not need to use the skill of closing a drawer if the drawer is already closed. Here, each skill is an MDP $\mathcal{M}^k = \langle \mathcal{E}, \mathcal{R}^k \rangle$, where the robot's environment $\mathcal{E}$ is shared across all tasks and skills, and $\mathcal{R}^k: \mathcal{S} \times \mathcal{A} \to \mathbb{R}$ is the reward function for the $k^{th}$ skill (e.g. the skill of opening a drawer).

\noindent \textcolor{rc}{\textbf{Problem Definition}}. \textcolor{rc}{Formally, $\acronym$ takes as input $N$ human-provided example images $\Psi^k = {s_1, \ldots, s_N}$ for each skill $k$ we intend to robot to learn. Our goal (i.e. $\acronym$'s output) is to learn the symbolic grounding of the pre- and post-conditions $g^k_{pre}, g^k_{post}: \mathcal{S} \rightarrow \{0, 1\}$, and a policy $\pi^k: \mathcal{S} \to \Pi(\mathcal{A})$ for each skill $k$, where $\Pi(\cdot)$ defines a probability distribution over a set. Each policy has a Q-function, $Q^k: \mathcal{S}\times\mathcal{A} \to \mathbb{R}$ for taking action $a$ from $s$ and following $\pi^k$ onward.}

\noindent \textbf{Variational Autoencoders (VAEs)}. VAEs compress high-dimensional observations $s$ such as images into an embedding $z$.
VAEs can be optimized by maximizing the evidence lower bound (ELBO): $\max_{p, q} \mathcal{L}_\text{vae}(s)$, where \inlineequation[eq:vae]{\mathcal{L}_\text{vae}(s) = \mathbb{E}_{q(z \mid s)} \left[\log p\left(s \mid z \right) \right] - D_\text{KL}\infdivx{q(z \mid s)}{p(z)}},
where $p$ denotes the generative model, and $q$ denotes the variational distribution.

\noindent \textbf{Symbolic planning}. A symbolic planner performs task planning in the symbolic domain. Concretely, given a set of candidate actions, each action's associated pre- and post-conditions, the current condition $h$ and a goal condition $g$, a symbolic planner outputs the sequence of actions that allows the robot to reach from $h$ to $g$. The pre-condition of an action is a set of predicates that defines the condition that needs to be satisfied in order for the action to be executable. The post-condition of a skill is a set of predicates that defines the effect of executing an action. For example, the pre-condition and post-condition of an ``Open left drawer'' action is ``Drawer is closed'' and ``Drawer is open'' respectively. In our context, the candidate actions are the primitive skills learned by $\acronym$.

\section{Example-Driven Model-Based RL ($\acronym$)}
\begin{figure}
    \centering
    \includegraphics[width=\textwidth]{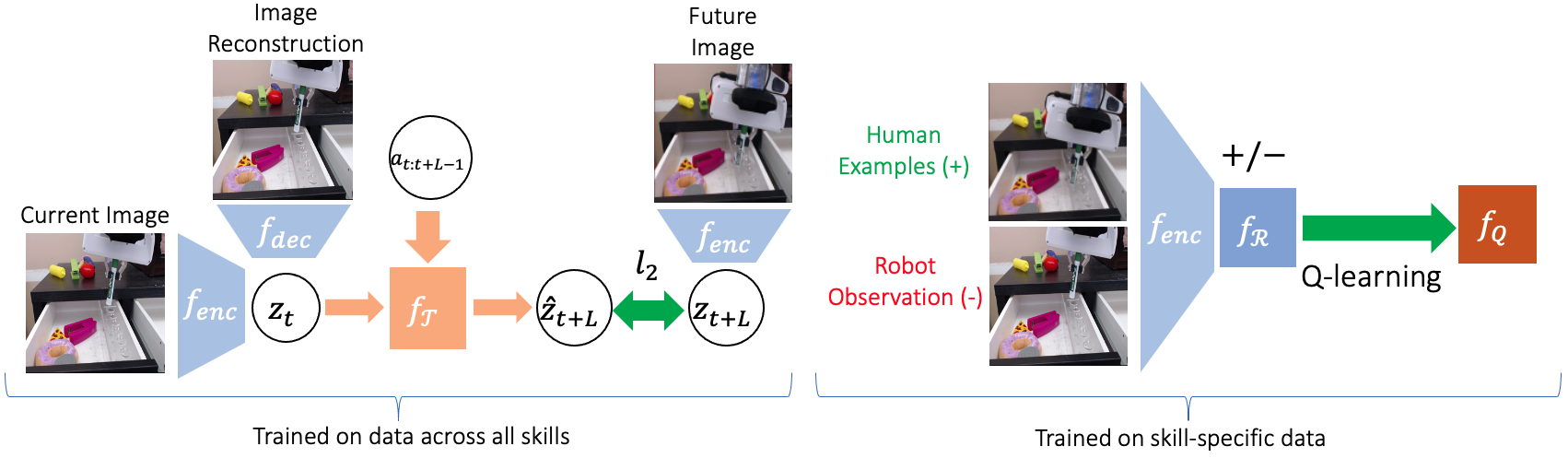}
    \caption{\small \textbf{Example-Driven Model-based RL ($\acronym$)}. 
    During training, $\acronym$ learns a VAE $\vae$ and a dynamics model $\dyn$ in the VAE latent space (\textbf{left}). For each skill, $\acronym$ uses human provided example images to train a success-classifier-based reward function $f_\mathcal{R}$, and associated Q function $f_Q$ (\textbf{right}). At test time actions are selected by model predictive control over the latent dynamics model using the Q-value as the planning cost. The reward function further serves as a success classifier that encourages the robot to reach a successful state, such as inserting the marker as shown in the figure.}
    \label{fig:ember}
    \vspace{-15pt}
\end{figure}
\textcolor{rc}{To learn a repertoire of skills that can be directly sequenced by an off-the-shelf symbolic planner to complete long-horizon visuomotor tasks, we propose Example-driven Model-Based RL, or $\acronym$. Below we first describe how $\acronym$ learns each visuomotor skill without any demonstration skills or trajectories during training, and then elaborate on how these learned skills can be directly sequenced to complete a long-horizon task at test time}. 

\subsection{\textcolor{rc}{Learning a Repertoire of Visuomotor Skills}}
During training, $\acronym$ learns a repertoire of low-level persistent visuomotor skills, such as opening a drawer or inserting a marker, using a set of example success images provided by a person for each task. The model learned by $\acronym$, denoted as $f$, is composed of four components: a global variational autoencoder (VAE) $\vae$, a global latent dynamics model $\dyn$, stage-specific binary reward functions $\rew$, and stage-specific Q-functions $\q$. The VAE and latent dynamics model jointly learn a low-dimensional representation of the image observations to address the challenge of efficiently learning many primitive skills; both are trained on data collected across all skills.
The skill-specific binary classifiers allow us to learn a reward function for each skill from human-provided images. 
We use the latent dynamics model to perform model predictive control with the skill-specific Q-function as the terminal planning objective.
The skill-specific Q-functions and reward functions are trained on skill-specific data. Next, we elaborate on each of these components.

\noindent \textbf{Learning a low-dimensional latent space with a VAE and a latent dynamics model}.
Performing a long-horizon task using a large number of visuomotor skills demands the algorithm to be sample-efficient and fast at execution time. To address both of these demands, we learn a single VAE $\vae = \{\enc, \dec\}$ to construct a latent space $\mathcal{Z}$. This VAE consists of an encoder $\enc: \mathcal{S} \rightarrow \Pi(\mathcal{Z})$ and a decoder $\dec: \mathcal{Z} \rightarrow \Pi(\mathcal{S})$. We train $\vae$ using data collected across all skills via Eq.~\ref{eq:vae}, where $p(s \mid z) \equiv \dec(s_t \mid z_t)$, $p(z) \sim \mathcal{N}(0, 1)$, and $q(z \mid s) \equiv \enc(z_t \mid s_t)$. By learning from all skills, the VAE can improve data efficiency. In practice, we use the greedy hierarchical VAE~\cite{wu2021greedy} as $\vae$ with the current image observation as a context frame. Based on this latent space, $\acronym$ learns a latent dynamics model $\dyn$ jointly with $\vae$ with data collected across all skills via $l_2$ loss:
\vspace{-5px}
\begin{equation}
     \min_{\vae, \dyn} \mathbb{E}_{s_{t:t+L}, a_{t:t+L} \sim \mathcal{D}}\bigg[ \sum_{l=1}^{L} -\mathcal{L}_\text{vae} (s_{t+l-1}) + \big({\dyn}(z_t, a_{t:t+l-1}) - z_{t+l} \big)^2\bigg]
     \label{eq:dyn}
\end{equation}
where $z_t \sim \enc(\cdot \mid s_t)$, $L$ is the rollout length of $\dyn$ and $\mathcal{D}$ is the dataset collected by the robot.

\noindent \textbf{Obtaining rewards by learning image classifiers}. Obtaining a reward function in RL typically requires some forms of human supervision such as reward engineering or demonstrations. In $\acronym$, binary image classifiers serve not only as reward functions for learning primitive visuomotor skills, but also as success detectors as well as grounding of raw pixels for long-horizon symbolic planning. Concretely, we learn a reward function $\rew$ for each skill $\mathcal{M}^k$ by training binary image classifiers to distinguish human-provided example images from robot-collected non-example images using the binary cross-entropy objective~\cite{fu2018variational,chen2020batch}. The reward learning objective for skill $k$ is: 
\begin{align}
    \max_{\rew} \mathbb{E}_{s^+ \sim \mathcal{D}^+, z^+ \sim \enc(\cdot \mid s^+)}\left[ \log\left( \rew(z^+)  \right) \right] + \mathbb{E}_{s^- \sim \mathcal{D}, z^+ \sim \enc(\cdot \mid s^+)}\left[\log\left( 1 - \rew(z^-)  \right) \right]
    \label{eq:rew}
\end{align}
Here, $\mathcal{D}^+$ is the dataset of images labeled as positive by a person (typically 100-200 images), $s^+$ are the images sampled from $\mathcal{D}^+$, and $s^-$ are sampled from all images in $\mathcal{D}$, the dataset collected by the robot, and are considered negative by default. In our experiments, these positive examples can be collected in 15-30 minutes per visuomotor skill, and 345 minutes in total for all 14 skills. \textcolor{rc}{No demonstration trajectories are collected.} 

At test time, the learned image classifiers are re-purposed as post-condition grounding of each skill for long-horizon task execution. Some of the pre-conditions the task planner requires can be derived from existing post-conditions. For example, for a drawer-opening skill, the pre-condition of ``drawer is not open'' can be derived from the post-condition of ``drawer is open'' by negating this post-condition. For the remaining pre-conditions that cannot be derived from any existing post-conditions, we train additional image classifiers using Eq.~\ref{eq:rew}.

\noindent \textbf{Learning Q-functions for model-based control}. 
While one can directly use the latent dynamics model for model-based planning, the difficulty of learning an accurate dynamics model for the full horizon of a primitive skill increases with the length of the skill itself. To alleviate this challenge and further accelerate skill learning, we learn a Q-function $\q$ in the latent space. Concretely, we perform Q-learning via the following objective for skill $k$:
\begin{align}
\min_{{\q}} \mathbb{E}_{s_t, a_t, s_{t+1} \sim \mathcal{D}}\left[\q \left(z_t, a_t \right) - \left(\overline{\rew}\left(z_{t+1}\right) + \gamma \overline{\rew} \left( z_{t+1} \right) \max_{a_{t+1}} \q(z_{t+1}, a_{t+1})\right)\right]^2\raisetag{20pt}
\label{eq:q}
\end{align}
where $\overline{\rew}(z) \equiv \mathds{1}\{\rew(z) > 0.5\}$, $z_t \sim \vae(\cdot \mid s_t)$, $z_{t+1} \sim \vae(\cdot \mid s_{t+1})$, and the Q-value target is computed by maximizing over $m_1$ randomly and uniformly sampled actions $a^{1:m_1}_{t+1}$. 

\setlength{\textfloatsep}{8pt} 
\begin{algorithm}
\small
\caption{$\acronym$ at Training Time: Learn a Repertoire of Visuomotor Skills}
\begin{algorithmic}[1]
  \State \textbf{Input}: Example images per skill; \textbf{Output}: $K$ primitive visuomotor skills
  \State $\mathcal{D} \leftarrow$ to empty dataset, $f \leftarrow$ random weights, $J \leftarrow$ gradient update steps, $L \leftarrow$ rollout horizon
  \For{Skill $k \in [1, K]$}
    \While{Not reaching target success rate for skill $k$ as determined by success classifier $f^k_{\mathcal{R}}$}{
        \State Collect trajectory $d=\{s_{1:H}, a_{1:H}\}$ using actions selected by Eq.~\ref{eq:test} and append to dataset $\mathcal{D}$
        \For{$j= 1 : J$}
        \State Sample a minibatch of trajectories
        of timesteps $L \leq H$:
        $\overline{\mathcal{D}}=\{s^{1:B}_{t:t+L}, a^{1:B}_{t:t + L}\} \sim \mathcal{D}$
        \State Update $\vae, \dyn$ jointly via Eq.~\ref{eq:dyn}, $\rew$ via Eq.~\ref{eq:rew}, and $\q$ via Eq.~\ref{eq:q}, all with $\overline{\mathcal{D}}$
        \EndFor
    }
    \EndWhile
\EndFor
\end{algorithmic}
\label{algo:EMBER-train}
\end{algorithm}
\vspace{-10px}
\begin{algorithm}[h]
\small
\caption{$\acronym$ at Test Time: Long-Horizon Task Execution via Skill Composition}
\begin{algorithmic}[1]
  \State \textbf{Input:} trained model $f$, symbolic goal condition $g$
  \State Compute current symbolic state $h$ using $f_{\mathcal{R}}$; $s \leftarrow \text{new image observation}$
  \While{$h \neq g$}
  \State Compute the first skill to perform $k \in [1, K]$ using symbolic planner given goal $g$ and current state $h$ 
  \While{$f^k_{\mathcal{R}}(\enc(s)) < 0.5$}
    \State Execute robot action $a$ via Eq.~\ref{eq:test} given $s$
    \State $s \leftarrow \text{current image observation}$
\EndWhile
\State Compute current symbolic condition $h$ from $f_{\mathcal{R}}$
\EndWhile
\end{algorithmic}
\label{algo:nsMORL-test}
\end{algorithm}
At both training and test time, we perform model-based planning in the latent space by first rolling out $\dyn$ for $L$ timesteps into the future and then evaluating the predicted future latent space observation at timestep $t+L$ using the learned Q-function, a process that is illustrated in Fig.~\ref{fig:ember}:
\begin{align}
    a_t = \left(\argmax_{a_{t:t+T-1}, a_{t+L} \in \mathcal{A}} \q\left(\dyn\left(z_t, a_{t:t+L-1}\right), a_{t+L}\right)\right)_1\text{, where }z_t \sim \vae(\cdot \mid s_t) 
    \label{eq:test}
\end{align}
Here, the outer and inner maximization operations are taken over $m_0$ and $m_1$ samples respectively.

Integrating these four components, $\acronym$ is trained simultaneously across all components from scratch (i.e. without pre-training) for learning a repertoire of visuomotor primitive skills (Alg.~\ref{algo:EMBER-train}). Concretely, the robot begins with an empty dataset $\mathcal{D}$ and can learn each skill in almost any order (Appendix~\ref{sec:reset}). During skill learning, the robot collects a new trajectory using actions selected from Eq.~\ref{eq:test} and updates all model components using $\mathcal{D}$. This process repeats until the current skill reaches a desirable success rate as computed by the success classifiers, before the robot moves on to learn the next skill. As the robot acquires more skills, the dataset $\mathcal{D}$ accumulates and improves the generalization of the visual dynamics model, which in-turn accelerates future skill learning. Wall-clock training time per skill and hyperparameters are detailed in Appendix~\ref{sec:training} and~\ref{sec:hyperparams} respectively.

\subsection{Composing Learned Visuomotor Primitive Skills for Long-Horizon Task Execution}
At test time, $\acronym$ provides the learned primitive skills and the symbolic grounding of each skill's post-conditions (in the form of success classifiers) to a high-level task planner (standard PDDL STRIPS symbolic planner~\cite{mcdermott1998pddl} in our case) to perform long-horizon task. As elaborated in Alg.~\ref{algo:nsMORL-test}, the human first specifies a long-horizon task using a symbolic goal condition $g$.
Next, $\acronym$ computes the current symbolic condition of the environment from raw pixel observations by passing the current image observations to each of the success classifiers.
Using both the symbolic goal condition and the current symbolic condition, the symbolic planner computes the sequence of skills to perform and executes the first skill of this sequence using actions selected from Eq.~\ref{eq:test}.
This procedure repeats until the current condition matches the goal condition, after which task execution terminates. Note that the arm position of the robot is reset after executing each skill. The predicate definitions, the pre- and post-conditions and the full details of the symbolic planner used in Section~\ref{exps} are in Appendix~\ref{appendix:symbols}. 

\section{Experiments}
\begin{figure}
    \centering
    \resizebox{\columnwidth}{!}{
    \subfloat[][Robot Environment]{
    \includegraphics[height=.28\textwidth]{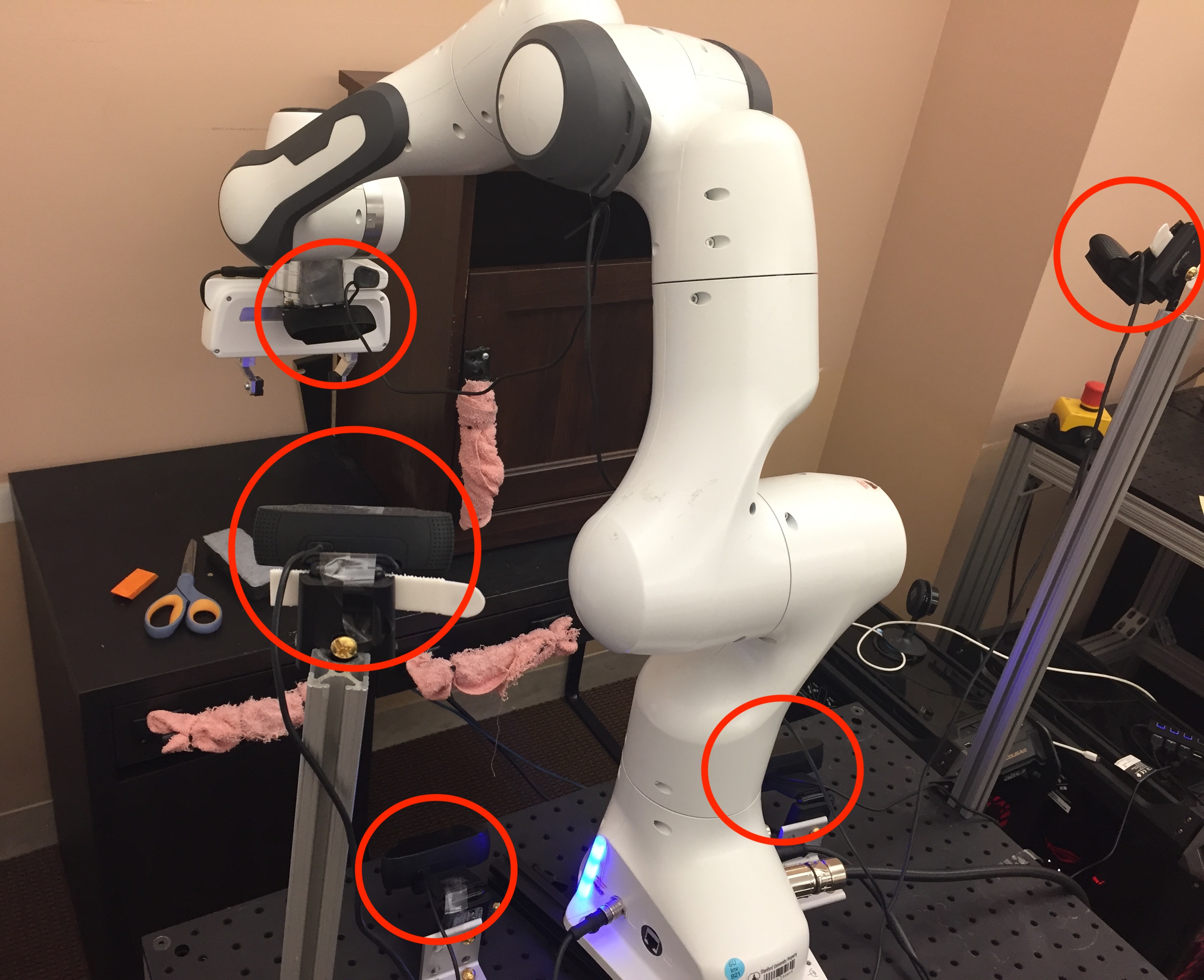}}
    \subfloat[][Train Objects]{
    \includegraphics[height=0.28\textwidth]{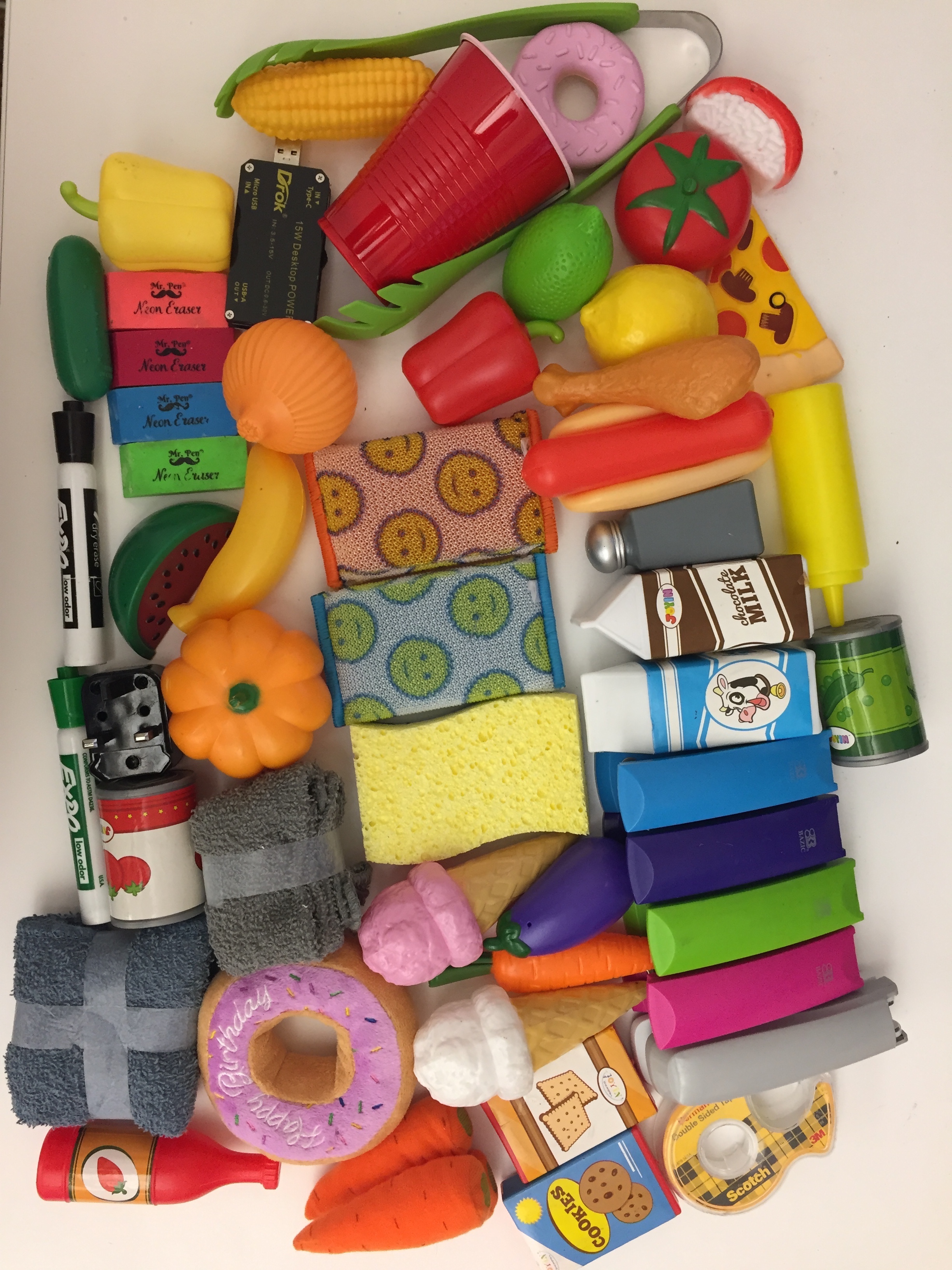}}
    \subfloat[][Validation Objects]{
    \includegraphics[width=0.28\textwidth,angle=90]{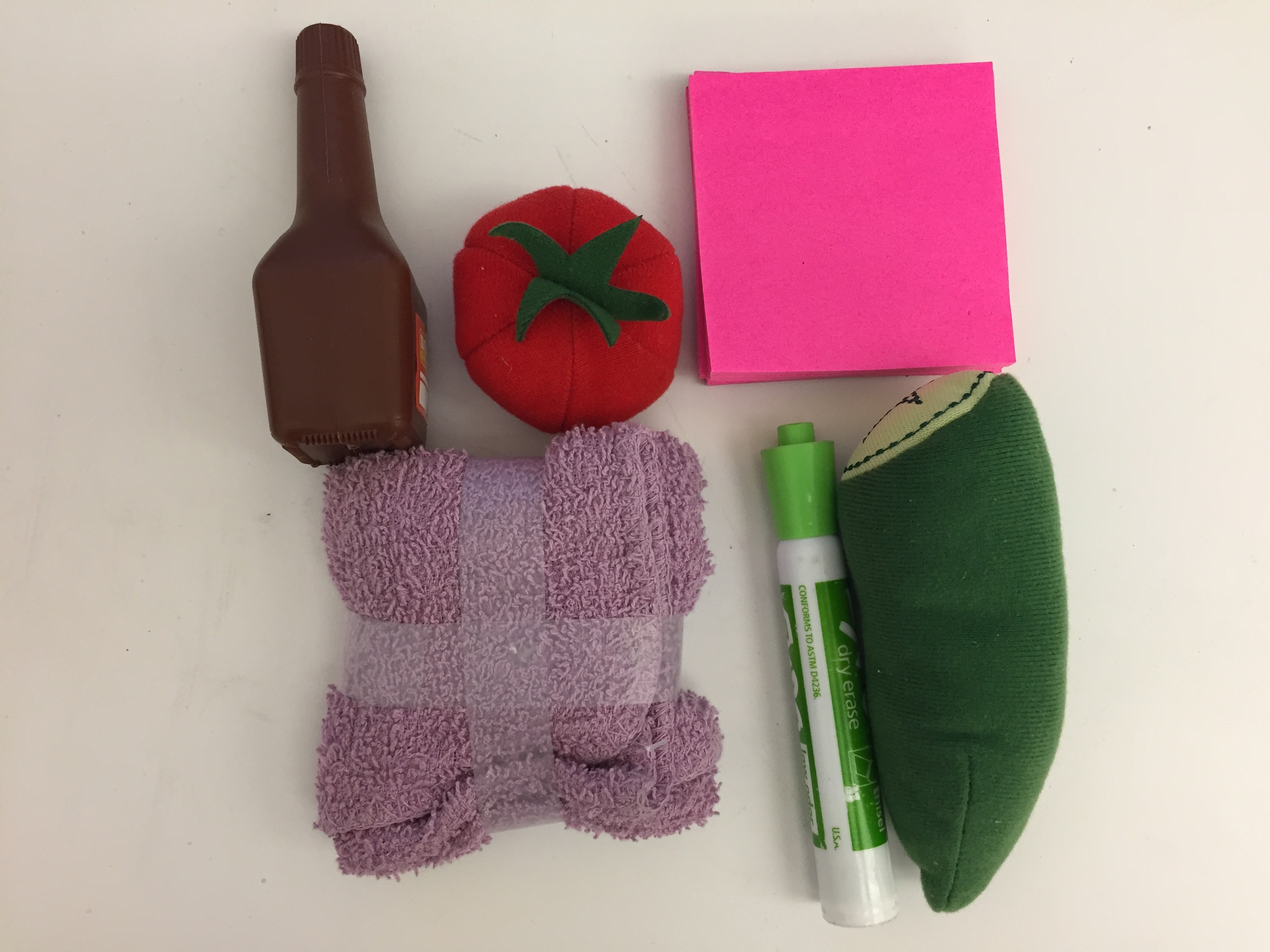}}
    \subfloat[][Test Objects]{
    \includegraphics[height=0.28\textwidth]{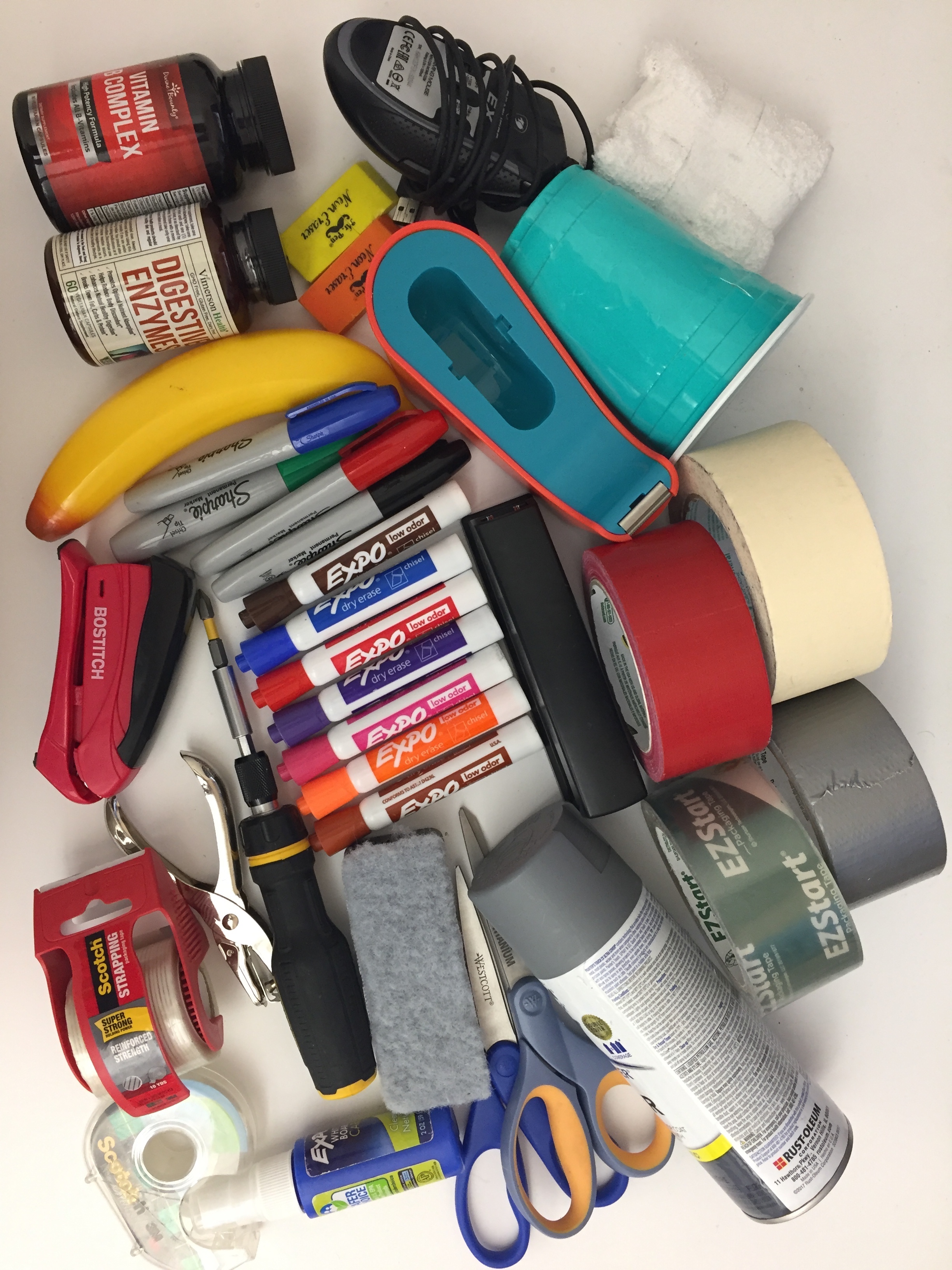}}
    \label{fig:novel}
    }
    \caption{\textbf{Real-Robot Experimental Setup}. In Fig.~\textbf{(a)}, a Franka Emika Panda robot is equipped with five RGB cameras (red circles) capturing 64$\times$64 RGB images. \textcolor{rc}{Quantitative justification for having five cameras is in Appendix~\ref{sec:ablation_camera}}. The robot interacts with a desk that has two drawers and a cabinet, as identified by three pink handles in Fig.~\textbf{(a)}. Fig.~\textbf{(b), (c) and (d)} visualize train, validation, and test objects used in all experiments.}
    \label{fig:objects}
    \vspace{-5pt}
\end{figure}
\label{exps}
Our experiments aim to answer four key questions: 1) how does $\acronym$ compare to prior RL algorithms on both performing primitive skills and when sequencing those skills to complete long-horizon tasks? 2) how necessary are stage-specific reward functions (as elaborated in Appendix~\ref{sec:stages}) compared to a single skill-specific reward function? 3) how important is replanning using the classifiers for the robustness of the algorithm? and 4) how important is each of the five camera observations to the performance of the primitive skills? 
To answer these questions, we conduct experiments across 14 primitive skills and three long-horizon tasks, using the experimental setup in Fig.~\ref{fig:objects}, in which a Franka Emika Panda robot has access to five cameras, each capturing 64$ \times$64 RGB images. One of the five cameras is mounted on the wrist. 
To test object generalization, all objects used in all evaluations are novel and unseen during training, except for the desk, drawers, cabinet, and the marker holder. Fig.~\ref{fig:objects} visualizes the train, validation, and test objects used in all experiments. See Appendix~\ref{appendix:exps} for experimental details. Project website at \url{https://sites.google.com/view/embr-site}.

\textbf{Comparisons.} We compare $\acronym$ to three prior methods:
\textbf{(1)} $\acronym$ w/o $\dyn$, which corresponds to $\acronym$ without the dynamics model, i.e. using only the Q-function and the VAE  for action selection; \textbf{(2)} $\acronym$ w/o $\vae$, $\dyn$ (Qt-Opt)~\cite{kalashnikov2018qt}, which is our re-implemented version of Qt-Opt~\cite{kalashnikov2018qt} with a VICE-like success classifier~\cite{fu2018variational} that does not learn a latent space and uses only the Q-function for action selection; and \textbf{(3)} BEE~\cite{chen2020batch}, which uses the success classifier instead of the Q-function for model predictive control.  All comparisons are representative of prior methods for vision-based robotic RL. For all experiments, the data used to train each method is the same dataset collected by $\acronym$ using Alg.~\ref{algo:EMBER-train}, which contains 300-400 trajectories collected in 16 hours per skill on average. This provides a more direct comparison and in effect makes the prior methods stronger since the challenge of online exploration is addressed by $\acronym$ during data collection.

\subsection{Short-horizon Primitive Skill Performance}
\begin{table*}[t]
\centering
\caption{\small Successful trials (out of 20) and success rates of visuomotor primitive skills. Here, $\acronym$ outperforms the three prior methods by 10-20\%, 15-25\%, and 50-65\% respectively on high-precision tasks.} 
\label{tab:short}
\small
\resizebox{0.98\columnwidth}{!}{
\begin{tabular}{c|l|c|c|c|c|c|c}
\toprule
\multirow{3}{*}{$k$} & \multirow{3}{*}{Primitive Skill} & \multirow{3}{*}{Precision} & \multirow{2}{*}{$\acronym$} & Prior Method & Prior Method 2: & Prior & Ablation 1: \\ 
& & & \multirow{2}{*}{(Ours)} & 1: $\acronym$ & $\acronym$ w/o $\vae$, & Method 3:& $\acronym$ w/ Skill- \\ 
& & & & w/o $f_{\mathcal{T}}$ & $\dyn$ (Qt-Opt)~\cite{kalashnikov2018qt} & BEE~\cite{chen2020batch} & specific Rewards\\ \midrule
1 & Insert marker into a marker holder & High & 20 (100\%) & 16 (80\%) & 15 (75\%) & 8 (40\%) & 20 (100\%) \\ 
2 & Grasp object from left drawer (single) & High & 18 (90\%) & 15 (75\%) & 14 (70\%) & 8 (40\%) & 15 (75\%) \\
& Grasp object from left drawer (clutter)  & High & 18 (90\%) & 14 (70\%) & 15 (75\%) &  7 (35\%) & 14 (70\%) \\ 
3 & Grasp object from right drawer (single) & High & 18 (90\%) & 16 (80\%) & 14 (70\%) & 5 (25\%) & 16 (80\%)\\ 
& Grasp object from right drawer (clutter) & High & 18 (90\%) & 15 (75\%) & 15 (75\%)  & 5 (25\%) & 16 (80\%) \\ 
4 & Grasp object from desk (single) & High& 18 (90\%) & 16 (80\%) & 15 (75\%) & 5 (25\%) & 12 (60\%)\\ 
& Grasp object from desk (clutter) & High & 18 (90\%) & 16 (80\%) & 15 (75\%) & 6 (30\%) & 12 (60\%)\\ 
5 & Open right drawer & Medium & 20 (100\%) & 18 (90\%) & 16 (80\%) & 14 (70\%) & 17 (85\%) \\ 
6 & Open left drawer & Medium &20 (100\%) & 19 (95\%) & 16 (80\%) & 8 (40\%) & 16 (80\%) \\ 
7 & Open cabinet & Medium & 20 (100\%) & 18 (90\%) & 17 (85\%) & 12 (60\%)  & 18 (90\%) \\ 
8 & Place object on desk & Low & 20 (100\%) & 20 (100\%) & 20 (100\%) & 20 (100\%) & 20 (100\%)\\
9 & Place object in cabinet & Low & 20 (100\%) & 20 (100\%) & 20 (100\%) & 20 (100\%) & 20 (100\%)\\
10 & Place object into left drawer & Low & 20 (100\%) & 20 (100\%) & 20 (100\%)  & 20 (100\%) & 20 (100\%) \\
11 & Place object into right drawer& Low  & 20 (100\%) & 20 (100\%) & 20 (100\%)  & 20 (100\%) & 20 (100\%) \\
12 & Close right drawer & Low & 20 (100\%) & 20 (100\%) & 20 (100\%) & 20 (100\%) & 20 (100\%)\\ 
13 & Close left drawer & Low & 20 (100\%) & 20 (100\%) & 20 (100\%) & 20 (100\%) & 20 (100\%) \\ 
14 & Close cabinet door & Low & 20 (100\%) & 20 (100\%) & 20 (100\%) & 20 (100\%) & 20 (100\%) \\ 
\bottomrule
\end{tabular}}
\vspace{-5px}
\end{table*}

To answer the first question, we compare all methods across 14 visuomotor skills. In Table~\ref{tab:short}, we observe that ``$\acronym$ (Ours)'' can reliably complete all 14 skills with at least 90\% success. Compared to ``$\acronym$ w/o $\dyn$'' and ``$\acronym$ w/o $\vae, \dyn$ (Qt-Opt)'', ``$\acronym$ (Ours)'' improves success rates by 10-20\% and 15-25\% across high-precision skills, 5-10\% and 15-20\% for medium-precision skills and 0\% for low-precision skills. We hypothesize that $\acronym$'s improvement over ``$\acronym$ w/o $\dyn$'' is due to its ability to alleviate overestimation of Q-values in the presence of limited training trajectories (300-400 per skill),
and the addition success rate improvement of $\acronym$ over ``$\acronym$ w/o $\vae, \dyn$'' is due to less overfitting of the success classifiers and Q-functions in the VAE latent space.
Finally, $\acronym$ outperforms BEE~\cite{chen2020batch} by 50-65\% on high-precision tasks because BEE~\cite{chen2020batch} cannot capture action consequences beyond its planning horizon and the model is not always accurate for the full horizon of a skill. 

\subsection{Long-horizon Task Performance}
\begin{figure}[t]
\vspace{-10pt}
    \centering
    \includegraphics[width=\textwidth]{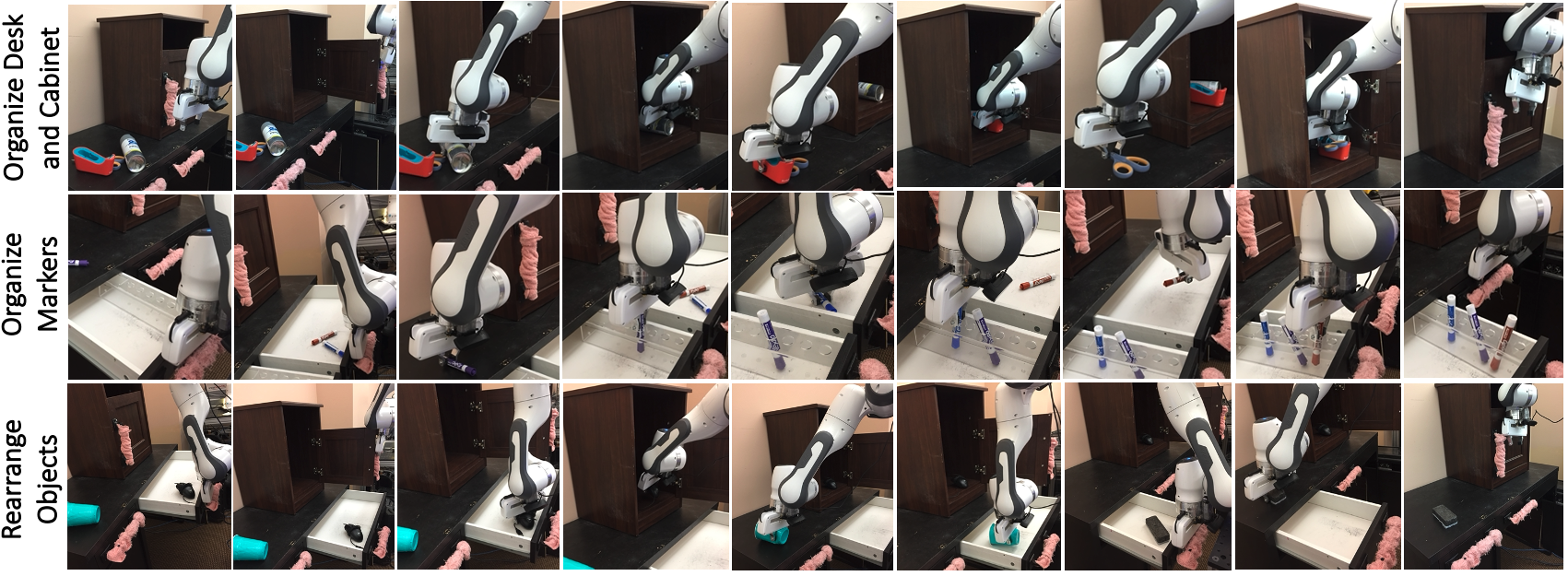}
    \caption{\small \textbf{Long-horizon Task Experiments}. $\acronym$ enables a Franka Emika robot to complete long-horizon manipulation tasks with novel objects from raw image observations. Here, the robot performs three long-horizon tasks with 85\% success, which involve a sequence of up to 12 unique primitive visuomotor skills. }
    \label{fig:long}
\end{figure}

\begin{table*}[t]
\centering
\caption{\small Successful trials (out of 20) and success rates of long-horizon tasks. $K^*$ and $K$ refer to the minimum number of skills that need to be sequenced and the minimum number of \textit{unique} skills required, respectively. Here, $\acronym$ outperforms the three prior methods by 15-25\%, 20-30\%, and 85\% respectively.} 
\label{tab:long}
\small
\resizebox{\columnwidth}{!}{
\begin{tabular}{c|l|c|c|c|c|c|c|c}
\toprule
& \multirow{3}{*}{Long-horizon Task} & \multirow{3}{*}{$K^*$} & \multirow{3}{*}{$K$} & \multirow{2}{*}{$\acronym$} & Prior Method & Prior Method 2: & Prior & Ablation 2: \\
&  & & & \multirow{2}{*}{(Ours)} &  1: $\acronym$ & $\acronym$ w/o $\vae$, & Method 3: & $\acronym$ \\ 
& & & & & w/o $f_{\mathcal{T}}$ & $\dyn$ (Qt-Opt)~\cite{kalashnikov2018qt} & BEE~\cite{chen2020batch} & w/o Replanning \\\midrule
1 & Organize Desk and Cabinet & 8 & 4 & 17 (85\%) & 14 (70\%) & 11 (55\%) & 0 (0\%) & 11 (55\%) \\ 
2 & Organize Markers & 9 & 5 & 17 (85\%) & 13 (65\%) & 11 (55\%) & 0 (0\%) & 10 (50\%) \\
3 & Rearrange Objects & 12 & 12 & 17 (85\%) & 15 (75\%) & 13 (65\%) & 0 (0\%) & 11 (55\%) \\
\bottomrule
\end{tabular}}
\end{table*}
Next, we compare $\acronym$ to the three prior methods when sequencing the learned visuomotor skills with symbolic planning to complete three challenging long-horizon tasks (Fig.~\ref{fig:long}): 
\vspace{-5px}
\begin{enumerate}[wide, labelindent=0pt]
\small
\item \textbf{Organize Desk and Cabinet (8 Skills)} - putting three novel objects cluttered on the desk into the cabinet. 
\item \textbf{Organize Markers (9 Skills)} - picking up three markers and inserting them into a holder.\footnote{\scriptsize In this task, we also equip all methods with a scripted skill that uses the edge of the desk as a supporting point to re-orient a marker.}
\item \textbf{Rearrange Objects (12 Skills)} - picking up three objects and placing them to the desk, cabinet or drawers. 
\vspace{-5px}
\end{enumerate}
Detailed task descriptions are in Appendix~\ref{appendix:tasks}. In Table~\ref{tab:long}, we find that ``$\acronym$ (Ours)'' completes these long-horizon visuomotor tasks with 85\% success, including the most challenging tasks that require up to 12 unique skills. Compared to ``$\acronym$ w/o $f_{\mathcal{T}}$'' and ``$\acronym$ w/o $\vae, \dyn$ (Qt-Opt)'', $\acronym$ leads to 10-20\% and 20-30\% improvement in success rates respectively because $\acronym$ achieves higher success rate on individual primitive skills that require higher precision, such as inserting marker and grasping object from desk or drawers. Finally, BEE~\cite{chen2020batch}, is unable to perform any long-horizon task due to low performance on skills with medium or high precision.
This reflects the importance of learning robust, high-success-rate primitive skills using $\acronym$.

\subsection{Ablations}
Next, we conduct six ablations to examine the importance of various components of $\acronym$. Ablation 3, which compares $\acronym$ with vs. without \textcolor{rc}{any one camera observation, is in Appendix~\ref{sec:ablation_camera}. Comparisons to the original BEE and training without stages are in Appendix~\ref{sec:originalbee} and~\ref{sec:scratch}.}

\textbf{Ablation 1: Stage-specific vs. skill-specific reward function}.
While $\acronym$ accelerates skill acquisition by learning a reward function for each stage of a skill, one can in principle learn just one reward function for each skill. In this ablation, we answer Question 2 by comparing $\acronym$ to an ablated $\acronym$ that learns each skill with a single skill-specific reward function. 
This does not affect data collection and exploration, which stage-specific rewards can also help.
In Table~\ref{tab:short}, we find that using stage-specific reward functions outperforms using a single skill-specific reward function by 10-30\% for higher precision skills such as grasping an object in various locations, indicating the importance of using stage-specific reward functions in long-horizon or higher precision skills.

\textbf{Ablation 2: Task performance with vs. without classifier-based replanning}.
While $\acronym$ improves the robustness of long-horizon task execution by using success classifiers during planning, one can in principle perform model-based planning without these classifiers. In this ablation, we answer Question 3 by comparing $\acronym$ to an ablation that executes each skill of a task sequentially without using the success classifiers to determine when the skill is successful. In Table~\ref{tab:long}, we find that long-horizon task execution with classifiers improves performance by 30-35\% success rate,
due to the robot's ability to detect success in a closed-loop manner until successful. 
These ablations and Table~\ref{tab:long} suggest that \emph{all} components of $\acronym$ are critical to completing long-horizon tasks.

\section{Conclusion and Future Work}
\textcolor{rc}{In this work, we propose $\acronym$: an example-driven model-based reinforcement learning algorithm for learning visuomotor skills that can be directly sequenced to perform long-horizon tasks from raw pixels.} Across three long-horizon visuomotor tasks which require up to 14 unique primitives and sequencing up to 12 skills, we find that $\acronym$ consistently achieves 85\% success rates, outperforms prior methods, and generalizes to novel objects. Nonetheless, $\acronym$ also has a number of important limitations, including the limited scope of tasks, the amount of supervision required to complete these complex long-horizon tasks, and the limited degree of generalization especially across different scenes. More specifically, the visuomotor skill definitions in $\acronym$ are currently object-specific, such as ``open left drawer'', which will scale poorly in the presence of many objects.  We discuss limitations in more detail in Appendix~\ref{sec:limitation}, each of which presents interesting directions for future work.
\newpage

\textbf{Acknowledgments}

The authors would like to thank members of the IRIS and RAIL labs for providing valuable feedback. Suraj Nair is funded in part by an NSF GRFP. This work was also supported in part by ONR grants N00014-20-1-2675 and N00014-21-1-2685.

\newpage
\appendix
\part{Appendix}
\parttoc
\section{Limitation and Future Work}
\label{sec:limitation}
While $\acronym$ significantly improves upon prior methods, there are a number of important limitations that we hope future work can address. 

\textbf{Task scope}. The current system does not learn skills that pay attention to object semantics. For example, the `grasp object from desk' skill involve grasping any object from the desk, rather than grasping a particular object of interest. Therefore, the current system is limited in the scope of long-horizon tasks that it can perform. Training skills that are semantic in nature is in principle possible with $\acronym$; doing so may require learning more powerful success classifiers for certain visuomotor skills, as they would need to pay attention to both spatial and semantic information.

\textbf{Amount of human supervision}. While the presented system was shown to complete 14 skills and 3 challenging long-horizon tasks, $\acronym$ completes tasks with a relatively large amount of human guidance: a human needs to determine the set of primitive skills, define the symbolic relationships, determine the stages for each primitive skill, and provide positive examples for each stage of each skill. Reducing the amount of human guidance without sacrificing on the complexity of skills that can be learned is an interesting avenue for future work.

\textbf{Environment generalization}. 
While $\acronym$ generalizes to new objects, it does not in its current form provide skills that generalize to new scenes or desks. This would likely require collecting data with greater variety in scenes and desks. Scaling data towards such generalization is also an interesting future direction.

\textbf{Planning}. 
While the technical focus of this work was to develop an RL algorithm that provides robust and grounded skills and evaluate how such skills can be sequenced, the symbolic planner employed in our system also has limitations. In particular, the symbolic planner we used does not handle partial observability. For example, the planner cannot tell whether an object is in the left drawer without opening the left drawer in the first place. We hope that future work can develop more sophisticated grounding mechanisms to handle such partial observability. 

\section{Clarification on $\acronym$'s Relationship with Hierarchical RL}
\textcolor{rc}{$\acronym$ is not a hierarchical RL algorithm because it does not learn a model over options, especially since in our case, the “model over options” is the off-the-shelf symbolic planner itself (i.e., the symbolic planner autonomously chooses which option to execute). Instead, we are learning a skill neural network for each of the 14 skills. These skills or options are not defined by the expert, but rather learned autonomously by using images of the terminal states as the only supervision signal. We recognize that different parts of the robotics and reinforcement learning community have different interpretations of the term ``model-based reinforcement learning''. In this paper, ``model-based RL'' only refers to learning each option efficiently by learning a visual dynamics model, rather than learning a high-level model over options.}

\section{Experimental Details}
\label{appendix:exps}
\subsection{Robot Observation and Action Space}
\textbf{Observation Space}. In all robotic experiments in this work, the robot is equipped with five $64 \times 64$ RGB cameras and no other observations. In practice, the robot's observation space has four out of five cameras: $\mathcal{S} = \mathbb{R}^{4 \times 64 \times 64 \times 3}$. 

\textbf{Action Space}. The robot's action space is a five-dimensional vector that specifies the translation and rotation of the Panda robot's end-effector as well as its gripper: $\mathcal{A} = \mathbb{R}^5$. The first three scalars in this vector denotes the $x, y, z$ translation of the end-effector in centimeters. The fourth scalar in this vector denotes the wrist rotation of the end-effector in radians. Finally, the five, last scalar in this vector is a binary scalar that commands the robot's gripper: 0 for fully opening the gripper, and 1 for fully closing the gripper. 

\subsection{Environment Reset}
\label{sec:reset}
Different from simulated learning, real-world robotic learning requires environment reset. In our experiments, we use the following reset strategies to reset each skill during data collection.

\textbf{Opening drawers}. For opening the left or right drawer, once the drawer is successfully opened, a forward pushing cartesian motion to re-close the drawer is sufficient to reset the environment.

\textbf{Closing drawers}. For closing the left or right drawer, once the drawer is successfully closed, a backward pulling cartesian motion to re-open the drawer is sufficient to reset the environment.

\textbf{Opening cabinet}. For opening the cabinet, once the drawer is successfully opened, a forward pushing cartesian motion to re-close the cabinet is sufficient to reset the environment.

\textbf{Closing cabinet}. For closing the cabinet, we use the open-cabinet skill to re-open the cabinet that has just been closed. Therefore, this skill should be learned after the ``Open cabinet'' skill is learned.

\textbf{Inserting marker}. For inserting a marker, once the marker is successfully inserted, an upward cartesian motion to pull-out a marker is sufficient to reset the environment. 

\textbf{Grasping object}. For grasping object skills, we throw a clutter of many training objects onto the desk or into the drawers. Once the grasp is successful, re-opening the closed gripper will be sufficient to reset the environment. In the case where the objects on the desk fall off the desk, a human picks them up once every few hours.

\textbf{Placing object}. For placing an object, re-closing the gripper and an upward cartesian motion to lift the object that has just been placed to the desk, drawers or cabinet is sufficient to reset the environment.

\begin{table*}[t]
\centering
\caption{\small Types and entities} 
\label{tab:types}
\small
\begin{tabular}{|r|l|c|}
\hline
& Types & Entities \\ \hline
1 & robot & robot \\ 
2 & location & desk, left-drawer, right-drawer, cabinet \\ 
3 & objects & object, marker \\ 
\hline
\end{tabular}
\end{table*}

\begin{table*}[t]
\centering
\caption{\small Predicates} 
\label{tab:predicates}
\small
\begin{tabular}{|r|l|c|}
\hline
& Predicates & Description \\ \hline
1 & (grasped robot) & robot's gripper is closed \\ 
2 & (at object location) & object is at location \\ 
3 & (opened location) & location is opened \\ 
4 & (in object robot) & object is in robot's gripper\\ 
5 & (inserted object) & object is inserted into the marker holder\\ 
\hline
\end{tabular}
\end{table*}

\subsection{Skills, Symbolic Actions, Predicates, Pre- and Post-conditions}
\label{appendix:symbols}
Types and entities defined symbolically are in Table~\ref{tab:types}. Predicates that build on these types are in Table~\ref{tab:predicates}. The mapping between the 14 learned primitive skills and their corresponding symbolic actions are in Table~\ref{tab:mapping}.

\subsection{Experimental Evaluation Protocol}
\subsubsection{Short-horizon Skill Evaluation}
For evaluation of primitive skills, all methods under comparison are given 100 timesteps to complete each skill. Inability to complete the skill under this timed budget denotes a failure and otherwise a success. The VAE and the latent dynamics model architecture we used is the GHVAE architecture~\cite{wu2021greedy}. The success classifiers and Q-functions are five-layer deep neural networks. 

\subsubsection{Long-horizon Task Evaluation}
At the beginning of each long-horizon task, all drawers and cabinet are closed. It is also required that all drawers and cabinets be closed at the end of each task to achieve task success. During evaluation, all methods are given $100 \times K^*$ timesteps to complete each long-horizon task, where $K^*$ is the minimum number of skills required to complete the task, as elaborated in Table~\ref{tab:long}. In effect, this is giving the robot 100 timesteps per skill that needs to be used in the task to complete the entire task. Inability to complete the skill under this timed budget denotes a task failure and otherwise a success.

\subsection{Explanation of Prior Method Comparison}
\label{appendix:baselines}
\subsubsection{Prior method 2: ``$\acronym$ w/o $\vae, \dyn$~\cite{kalashnikov2018qt}''}
We note that while we name the second prior method ``$\acronym$ w/o $\vae, \dyn$~\cite{kalashnikov2018qt}'' by citing Qt-Opt~\cite{kalashnikov2018qt}, this prior method is similar to Qt-Opt except in the following ways:
\begin{enumerate}
    \item Qt-Opt uses high-resolution images $472 \times 472$ as image observation while $\acronym$ use $64 \times 64$ images
    \item Qt-Opt uses distributed machine learning mechanisms while ``$\acronym$ w/o $\vae, \dyn$~\cite{kalashnikov2018qt}'' does not use distributed mechanisms
    \item ``$\acronym$ w/o $\vae, \dyn$~\cite{kalashnikov2018qt}'' uses only positive example images and does not use negative example images
    \item The original Qt-Opt algorithm~\cite{kalashnikov2018qt} uses a specialized CNN architecture, while ``$\acronym$ w/o $\vae, \dyn$~\cite{kalashnikov2018qt}'' uses the ``ResNet152'' architecture in both the success classifier and the Q-function, which is the largest CNN we can fit into our computational resources. 
\end{enumerate}

\subsubsection{Prior method 3: ``BEE~\cite{chen2020batch}''}
We also note that for the third prior method ``BEE~\cite{chen2020batch}'', we use the same rollout timesteps of $L=5$ as in $\acronym$ for fair comparison.

\subsection{Short-horizon Skill Descriptions}
\label{appendix:skills}

\textbf{Insert a marker into a marker holder}. In this skill, the robot needs to insert a marker of unseen texture or geometry (while the exact geometry is unseen and novel, markers in general are quite geometrically similar) into a marker holder. At the beginning of the skill, the marker is assumed to be in the robot's gripper and oriented vertically. The skill is considered a success when the marker is successfully inserted into the marker holder and a failure otherwise. 

Before the insert-marker skill is executed, we also equip the robot with an additional scripted skill of changing the orientation of the marker from horizontal to vertical. This skill uses the edge of the desk as a supporting point to turn a marker of any texture (since it is scripted) from horizontal to vertical. While one can also acquire this skill using $\acronym$, we found that the scripted version of the skill performs best primarily due to the high force-torque intensity of this skill.

\textbf{Grasp object from left drawer (single)}. In this skill, the robot needs to pick-up a single object of novel geometry and/or texture from the left drawer. At the beginning of each trial, the object is already in the left drawer. The initial position is randomized but not at the corners of the left drawer, in which case grasping is not mechanically possible using the Franka Emika Panda parallel-jaw gripper. The initial orientation of each object is also randomized. The skill is considered a success when the object is successfully picked up and a failure otherwise.

\textbf{Grasp object from left drawer (clutter)}. In this skill, the robot needs to pick-up a single object from a three-object clutter, each with novel geometry and/or texture, in the left drawer. At the beginning of each trial, three novel objects are already in the left drawer. The initial position of each object is randomized but not at the corners of the left drawer, in which case grasping is not mechanically possible using the Franka Emika Panda parallel-jaw gripper. The initial orientation of each object is also randomized. The skill is considered a success when any one object is successfully picked up and a failure otherwise.

\textbf{Grasp object from right drawer (single)}. In this skill, the robot needs to pick-up a single object of novel geometry and/or texture from the right drawer. At the beginning of each trial, the object is already in the right drawer. The initial position is randomized but not at the corners of the right drawer, in which case grasping is not mechanically possible using the Franka Emika Panda parallel-jaw gripper. The initial orientation of each object is also randomized. The skill is considered a success when the object is successfully picked up and a failure otherwise.

\textbf{Grasp object from right drawer (clutter)}. In this skill, the robot needs to pick-up a single object from a three-object clutter, each with novel geometry and/or texture, in the right drawer. At the beginning of each trial, three novel objects are already in the right drawer. The initial position of each object is randomized but not at the corners of the right drawer, in which case grasping is not mechanically possible using the Franka Emika Panda parallel-jaw gripper. The initial orientation of each object is also randomized. The skill is considered a success when any one object is successfully picked up and a failure otherwise.

\textbf{Grasp object from desk (single)}. In this skill, the robot needs to pick-up a single object of novel geometry and/or texture from the desk. At the beginning of each trial, the object is already on the desk. The initial position and orientation of the object is randomized. The skill is considered a success when the object is successfully picked up and a failure otherwise. 

\textbf{Grasp object from desk (clutter)}. In this skill, the robot needs to pick-up a single object from a three-object clutter, each with novel geometry and/or texture,  from the desk. At the beginning of each trial, the object is already on the desk. The initial position and orientation of the object is randomized. The skill is considered a success when the object is successfully picked up and a failure otherwise. If any of the three objects on the desk is pushed off the desk, the episode is also considered a failure.

\textbf{Place an object into left drawer}. In this skill, the robot needs to place a single object of novel geometry and/or texture into the left drawer. In the beginning of each trial, the object is already in the gripper and the left drawer is already open. The skill is considered a success when the object is successfully placed into the left drawer and a failure otherwise. 

\textbf{Place an object into right drawer}. In this skill, the robot needs to place a single object of novel geometry and/or texture into the right drawer. In the beginning of each trial, the object is already in the gripper and the right drawer is already open. The skill is considered a success when the object is successfully placed into the right drawer and a failure otherwise. 

\textbf{Place an object on desk}. In this skill, the robot needs to place a single object of novel geometry and/or texture on the desk. In the beginning of each trial, the object is already in the gripper. The skill is considered a success when the object is successfully placed on the desk and a failure otherwise. 

\textbf{Place an object in cabinet}. In this skill, the robot needs to place a single object of novel geometry and/or texture into the cabinet. In the beginning of each trial, the object is already in the gripper and the cabinet is already open. The skill is considered a success when the object is successfully placed in the cabinet and a failure otherwise. 

\subsection{Long-horizon Task Descriptions}
\label{appendix:tasks}
\noindent \textbf{Organize Desk and Cabinet}. In this task, the robot needs to pick-up all three novel objects (unseen during training) cluttered on the desk in any order and place them in the cabinet. After placing, the robot is also required to close all drawers and cabinet in the end. In this task, there will always be three cluttered objects on the desk unseen during training, and the position and orientation of each object is non-deterministic. All three cluttered objects have novel geometries and/or textures that are unseen during training. These novel objects were sampled randomly from Fig.~\ref{exps} (d). While one can test the algorithm across a flexible number of objects at the desk, we fix the number of objects per experiment to three, considering that the overall task performance depends on the total number of skills required to complete the entire task. Some of these objects are also semi-transparent, such as the scotch tape. The task is only considered successful when all three objects are successfully placed into the cabinet and the cabinet and the drawers are closed. 

\noindent \textbf{Organize Markers}. In this task, the robot needs to pick-up three unseen markers (unseen texture, seen geometry) and insert it into a marker holder in the left drawer. All three markers have novel textures that are unseen during training. The marker holder is not mounted to the drawer and can move freely in the left drawer. Each marker can either be on the desk or in the right drawer. Although we could have also placed some markers in the left drawer to make the task more difficult, a marker in the left drawer is not mechanically graspable by the Frank Emika robot gripper once the marker holder is placed in the same drawer (the left drawer will not be wide enough for the gripper to grasp anything). For example, at the beginning of a trial, two markers can be randomly placed on the desk (random position and orientation), and one marker can be randomly placed in the right drawer (random position and orientation). 

To re-orient each marker, we also equip all methods with a scripted skill that uses the edge of the desk as a supporting point to re-orient the marker in the gripper. This skill is executed right before executing the "Insert marker into a marker holder'' skill. We emphasize that only this one skill is scripted in the robot's skill repertoire. All other 14 skills in the robot's skill repertoire is both visuomotor and learned.

\noindent \textbf{Rearrange Objects}. In this task, the robot needs to rearrange three objects to their respective desired locations. All three objects have novel geometries and/or textures that are unseen during training. Initially, each of the three novel objects will be placed on the desk (object1), in the left drawer (object2), and in the right drawer (object3) respectively. We evenly place these objects to maximize the number of unique visuomotor skills the robot needs to use in order to complete the task. In other words, there will always be one object on the desk, one object in the left drawer, and one object in the right drawer, so that the robot needs to use all three visuomotor skills to complete the entire long-horizon task: ''Grasp object from left drawer'', ``Grasp object from right drawer'', and ``Grasp object from desk''. However, the identity of each novel object is random, so are their positions and orientations.

\subsection{Explanation of Ablation Experiments}
\textbf{Ablation 1:  Stage-specific vs.  skill-specific reward function}
In this ablation, we re-train $\acronym$'s Q-functions using just one stage per skill using the same dataset $\mathcal{D}$ for fair comparison. Therefore, for skills that only have one stage, the success rates would be the same for using stage-specific and  skill-specific reward functions.

\textbf{Ablation 2:  Task performance with vs.  without classifier-based replanning}. In this ablation, we disable classifier-based replanning and thus didn't need to re-train any of $\acronym$'s neural networks, i.e. the VAE, the dynamics model, the success classifiers, and the Q-functions. Concretely, we execute each skill for 100 timesteps without early termination and move on to the next skill without success detection. Notice that in this case, the success classifiers are still used for training the Q-function; they are just not used for replanning during long-horizon task execution.

\section{Additional Ablations: $\acronym$ vs. Original BEE}
\label{sec:originalbee}
\textcolor{rc}{The only changes to the BEE comparison are to give it the same architecture as $\acronym$, specifically to make the comparison of algorithms more fair. We compared to the original BEE algorithm using the code from \url{https://github.com/stanford-iris-lab/batch-exploration}. For all grasping skills, we observe that the original BEE achieves 0\% success rate. This shouldn’t be surprising at all: the VAE in BEE is much smaller than the version of BEE we compared to in our pre-revision paper. The VAE learned from the original BEE algorithm cannot reconstruct any details of graspable objects, which makes the original BEE perform much worse than either our re-implemented version of BEE in our pre-revision paper, or the $\acronym$ algorithm. }
\begin{table*}[t]
\centering
\caption{\small Ablation 4: $\acronym$ without Stages ($\acronym$ Data) vs. $\acronym$ without Stages (Recollect Data). Successful trials (out of 20) and success rates of visuomotor primitive skills.} 
\label{tab:scratch}
\small
\resizebox{0.98\columnwidth}{!}{
\begin{tabular}{c|l|c|c|c|c}
\toprule
$K$ & Primitive Skill & Precision & $\acronym$ & $\acronym$ w/o Stages ($\acronym$ Data) & $\acronym$ w/o Stages (Recollect Data) \\  \midrule
1 & Insert marker into a marker holder & High & 20 (100\%) & 20 (100\%) & \textcolor{rc}{20 (100\%)} \\ 
5 & Open right drawer & Medium &20 (100\%) &17 (85\%) & \textcolor{rc}{0 (0\%)}\\
6 & Open left drawer & Medium &20 (100\%) &16 (80\%) & \textcolor{rc}{0 (0\%)}\\
7 & Open cabinet & Medium & 20 (100\%) &18 (90\%) & \textcolor{rc}{0 (0\%)}\\
11 & Place object into left drawer & Low & 20 (100\%) & 20 (100\%) &\textcolor{rc}{20 (100\%)}\\
12 & Place object into right drawer& Low  & 20 (100\%) & 20 (100\%) &\textcolor{rc}{20 (100\%)}\\
13 & Place object on desk & Low & 20 (100\%) &20 (100\%)&\textcolor{rc}{20 (100\%)}\\
14 & Place object in cabinet & Low & 20 (100\%) &20 (100\%)&\textcolor{rc}{20 (100\%)}\\
\bottomrule
\end{tabular}}
\vspace{-5px}
\end{table*}

\section{Additional Ablations: $\acronym$ with Stage-Specific Rewards vs. $\acronym$ with Skill-Specific Rewards Trained from Scratch}
\label{sec:scratch}
\textcolor{rc}{In addition to comparing to $\acronym$ w/o Stages ($\acronym$ Data) (i.e. $\acronym$ without Stage-specific Rewards but using data already collected from the original $\acronym$ algorithm), we also compared to $\acronym$ without Stages (Recollect Data) (i.e. an additional ablated version of ``$\acronym$ without Stage-specific Rewards'' that collects data from scratch). In Table~\ref{tab:scratch}, we observe that ``$\acronym$ w/o Stages (Recollect Data)'' cannot explore medium-precision skills that have multiple stages. This includes all cabinet and drawer-opening skills. For skills that only have one stage, the performance is the same as ``$\acronym$ w/o Stages ($\acronym$ data)''.}

\begin{table*}[t]
\centering
\caption{\small Ablation 3: Skill performance with vs. without any of the five cameras. Successful trials (out of 20) and success rates of visuomotor primitive skills. } 
\label{tab:camera}
\small
\resizebox{\columnwidth}{!}{
\begin{tabular}{|r|l|c|c|}
\hline
$K$ & Primitive Skill & $\acronym$ With Wrist Camera & $\acronym$ Without Wrist Camera \\ \hline
2 & Grasp object from left drawer (single) & 18 (90\%) & 14 (70\%)\\
& Grasp object from left drawer (clutter)  & 18 (90\%) & 15 (75\%) \\ 
3 & Grasp object from right drawer (single) & 18 (90\%) & 12 (60\%)\\
& Grasp object from right drawer (clutter)  & 18 (90\%) & 14 (70\%) \\ 
4 & Grasp object from desk (single) & 18 (90\%) & 12 (60\%)\\ 
& Grasp object from desk (clutter) & 18 (90\%) & 13 (65\%) \\ \hline
$K$ & Primitive Skill & $\acronym$ With Left Shoulder Camera & $\acronym$ Without Left Shoulder Camera \\ \hline
7 & Open Cabinet & \textcolor{rc}{20 (100\%)} & \textcolor{rc}{3 (15\%)}\\\hline
$K$ & Primitive Skill & $\acronym$ With Right Shoulder Camera & $\acronym$ Without Right Shoulder Camera \\\hline
5 & Open Right Drawer & \textcolor{rc}{20 (100\%)} & \textcolor{rc}{11 (55\%)}\\
6 & Open Left Drawer & \textcolor{rc}{20 (100\%)} & \textcolor{rc}{13 (65\%)}\\\hline
$K$ & Primitive Skill & $\acronym$ With Left Waist Camera & $\acronym$ Without Left Waist Camera \\ \hline
2 & Grasp object from left drawer (single) & \textcolor{rc}{18 (90\%)} & \textcolor{rc}{11 (55\%)} \\\hline
$K$ & Primitive Skill & $\acronym$ With Right Waist Camera & $\acronym$ Without Right Waist Camera \\ \hline
3 & Grasp object from right drawer (single) & \textcolor{rc}{18 (90\%)} & \textcolor{rc}{6 (30\%)} \\\hline
\end{tabular}}
\end{table*}
\section{Additional Ablations: Task Performance With vs. Without Any of the Five Cameras}
\label{sec:ablation_camera}
\textcolor{rc}{In our experiments, the robot has access to RGB images from five cameras, one of which is mounted on the wrist of the robot. To measure the contribution of each camera to the observed performance, we re-trained $\acronym$'s success classifier and Q-function using only four of the five cameras, excluding the wrist, left-shoulder, right-shoulder, left-waist or the right-waist camera. Note that in all ablations, we exclude the particular camera observation from the robot's observation space and then re-train $\acronym$, using the same dataset $\mathcal{D}$, for fair comparison. In Table~\ref{tab:camera}, we found that each of the five cameras is important to the overall performance in that it improves the success rate of specific skills by 15-85\%.}

\textcolor{rc}{The reliance on a large set of perception systems is mainly the result of the narrow field of view (70 degrees) of each of the five cameras, such that:
\begin{enumerate}
    \item Only the left shoulder camera can see the entire cabinet 
    \item Only the right shoulder camera can see the entire left and right drawers when the robot is pulling the drawer handle simultaneously - this is especially important during our long-horizon task experiments since the robot needs to know whether both drawers are open or closed.
    \item Only the left waist camera can clearly see the entire marker holder for “insert marker” skills and all individual objects in the left drawer
    \item Only the right waist camera can clearly see all individual objects in the right drawer
\end{enumerate}
The reason why as many as five cameras are needed is because there are 14 different skills with varying degrees of precision (from inserting a marker to opening a cabinet on the other end of the desk). Learning to do all of 14 skills requires a global view of the scene. The $\acronym$ algorithm is compatible with an arbitrary number of cameras, and in principle, a single camera with a sufficient field-of-view would work as well. Using five narrow-field-of-view cameras in our particular office desk setup doesn’t affect the applicability, transfer, and replicability of the $\acronym$ framework, because $\acronym$ is fully flexible and compatible with an arbitrary number of cameras, and one can add or remove cameras as they wish. }
\subsection{Excluding Wrist Camera}
\textcolor{rc}{In Table~\ref{tab:camera}, we find that including the wrist camera observation leads to a 15-30\% improvement in grasping skill success rate, highlighting the importance of the wrist camera observation.}

\subsection{Excluding Left-Shoulder Camera}
\textcolor{rc}{In Table~\ref{tab:camera}, we find that including the left-shoulder camera observation leads to a 85\% improvement in the success rate of opening the cabinet on the office desk, because the left-shoulder camera is the only camera that can fully observe the cabinet.}

\subsection{Excluding Left-Shoulder Camera}
\textcolor{rc}{In Table~\ref{tab:camera}, we find that including the right-shoulder camera observation leads to a 35-45\% improvement in the success rate of opening the drawers of the office desk, because the right-shoulder camera is the only camera that can see the entire left and right drawer when the robot is pulling the drawer handle.}

\subsection{Excluding Left-Waist Camera} \textcolor{rc}{In Table~\ref{tab:camera}, we find that including the left-waist camera observation leads to a 35\% improvement in the success rate of grasping an object from the left drawer, because only the left-waist camera can clearly see individual objects in the left drawer.}

\subsection{Excluding Right-Waist Camera} \textcolor{rc}{In Table~\ref{tab:camera}, we find that including the right-waist camera observation leads to a 60\% improvement in the success rate of grasping an object from the right drawer, because only the right-waist camera can clearly see individual objects in the right drawer.}

\section{Amount of Human Supervision and Robot Training Required}
\label{sec:training}
For each stage of a skill, providing a set of example images typically takes 15 minutes. Providing a set of example images for a skill takes $15 \times N_k$ minutes, where $N_k$ is the total number of stages for the skill.
The amount of robot training time is 16 hours on average per skill, regardless of how many stages the skill has. The number of trajectories per skill is 300-400 trajectories on average.

\section{Hyperparameters}
\label{sec:hyperparams}
Table~\ref{tbl:hyperparams} details the hyperparameters for $\acronym$.
\begin{table}[H]
    \centering
    \caption{Hyperparameters}
    \resizebox{0.8\linewidth}{!}{
    \begin{tabular}{l|l|c}
    Category & Hyper-parameter & Value \\ \hline
    Train Only & Episode Length / Horizon $H$ & 50 \\\hline
    \multirow{2}{*}{Test Only} & \# rollout trajectories $m_0$ & 150 \\
    & Episode Length / Horizon $H$ & 100 \\
     \hline
    \multirow{4}{*}{Train and Test} & Discount Rate $\gamma$ & 0.98 \\
    & Rollout Horizon $L$ & 5\\
    & Number of action samples during Q-maximization $m_1$ & 200 \\
    & Number of Camera Observations & 4 \\
    \end{tabular}
    }
    \label{tbl:hyperparams}
\end{table}

\end{document}